# How Usable is Automated Feature Engineering for Tabular Data?

Bastian Schäfer[1]  Lennart Purucker[1]  Maciej Janowski[1, 2]  Frank Hutter[4, 3, 1]

[1]University of Freiburg
[2]University of Technology Nuremberg
[3]ELLIS Institute Tübingen
[4]Prior Labs

**Abstract**  Tabular data, consisting of rows and columns, is omnipresent across various machine learning applications. Each column represents a feature, and features can be combined or transformed to create new, more informative features. Such feature engineering is essential to achieve peak performance in machine learning. Since manual feature engineering is expensive and time-consuming, a substantial effort has been put into automating it. Yet, existing automated feature engineering (AutoFE) methods have never been investigated regarding their usability for practitioners. Thus, we investigated 53 AutoFE methods. We found that these methods are, in general, hard to use, lack documentation, and have no active communities. Furthermore, no method allows users to set time and memory constraints, which we see as a necessity for usable automation. Our survey highlights the need for future work on usable, well-engineered AutoFE methods.

## 1 Introduction

High-performing machine learning for tabular data relies on effective feature engineering [1]. Therefore, practitioners manually engineer new features by combining existing features [2]. Performing feature engineering by hand is a time-consuming process that is challenging and requires immense expertise [3, 4, 5]. Thus, automated feature engineering (AutoFE) is emerging as a replacement to manually feature engineering [6, 7].

Despite the potential of AutoFE for practitioners, there exists no guidance for newcomers on which AutoFE method to use. Practitioners have no overview of how easy it is to use AutoFE and whether it makes feature engineering less time-consuming. Thus, we set out to manually investigate the usability of AutoFE. Our investigation enables practitioners to determine whether to use AutoFE methods and highlights problems in current methods for the research community.

We investigated 53 AutoFE methods and evaluated their usability with respect to nine criteria ranging from code and documentation quality to usability. Our results demonstrate that the majority of AutoFE methods are not usable for practitioners. Additionally, none of the methods are usable in a fully automated context without issues.

**Contribution**. We provide an overview of the usability of existing AutoFE methods with a closer look at the code, the documentation, the open-source community, and support for user constraints. Additionally, we analyze whether a method is inductive or not. Our results clearly show that most AutoFE methods are hard to use and that no method is usable in a real-world AutoML context.

## 2 Related Work

**Surveys or Benchmarks**. Existing benchmarks lack a complete picture of state-of-the-art AutoFE methods. Some important methods are not considered, as the benchmarks are tailored to particular applications of feature engineering or address only feature selection methods cf. [8, 9, 10, 11].

**AutoFE Method.** The traditional approach in AutoFE is the black-box-approach and consists of two steps: At first, the generation of new features, and secondly, the reduction of the pool of candidate features by selecting the most promising ones. The selected features are then added to the dataset and used for the machine learning task [12, 13, 14]. Other popular approaches leverage reinforcement learning, which trains an agent to select the best features [15, 16, 17], or utilize the context and world knowledge of large language models [5, 18].

## 3 Our Survey

We investigated 53 AutoFE methods we found in the literature. For each method, we **(1)** search for open-source code; **(2)** evaluate the available code and API quality in a *subjective* manner, considering criteria for successful open-source libraries [19, 20, 21, 22]; **(3)** search for the documentation of the code; **(4)** investigated whether the method has an active community by determining the last update of the method and user engagement on public repositories; **(5)** check if a user could pass constraints, such as a time or memory limit, to the method; and **(6)** we investigate if the method is built for inductive or transductive learning [1].

With inductive learning, we refer to a method that does not look at the test samples during training, as is the standard assumption for machine learning on tabular data. Transductive learning refers to a method that incorporates information from the test samples (without the labels) to influence feature engineering. We highlight this difference as we observed that the AutoFE community often overlooked the positive impact of transductive learning in benchmarks, even in prominent work such as OpenFE [12]. That is, many papers incorrectly compare transductive methods to inductive methods, ignoring that the two scenarios are not directly comparable.

We test the usability of all AutoFE methods by attempting to download, install, and then execute them in a test pipeline. The test pipeline tries to run the AutoFE method once on datasets from the AutoML benchmark [23], and verifies that the method returns data that can be passed (after reformatting) to a tabular machine learning model.

Lastly, we conclude whether we consider a method to be an AutoFE method. We considered a method to be AutoFE if we were able to use its code with minimal effort and constrain its time and memory budget. A method can only be AutoFE if it is easy to use, thereby reducing time-consuming processes and not requiring immense expertise. Moreover, a method must be enable users to constrain the resource usage to be considered fully automatic [24]. Feature engineering is particularly expensive in terms of memory. Hence, we require both time and memory constraints. In our study, we only investigated open-source methods. We would have also investigated closed-source methods for their usability, but all closed-source methods we found did not provide any usable (API) interface. Instead, they were academic papers without code. We provide an exhaustive description of our investigation in the appendix.

## 4 Results

We investigated a total of 53 AutoFE methods. 26 methods do not have code. 27 methods have code, whereby for all but four methods, the code was provided together with the paper. For the other four methods, we found an unofficial implementation, that is, code written by someone other than the authors or the organization behind the paper. From the 27 methods with code, we were only able to run and use 16 within our test pipeline. Thus, in the end, a total of 43 methods we found in the literature are not usable at all. For the remaining, we investigated how usable the methods are. Table 1 provides a summary of our investigation, while we give the complete overview of our investigation in Section 5.

We observe that not even half of the methods are open-source, only 10 of them provide proper documentation, and only 7 have a community continuously developing the project. The degree of full automation is zero for all of the methods; none enable users to a time and memory limit constraint. Of the 27 open-source methods, 20 prevent rely only on inductive learning.

Table 1: Summary of usability, code availability, existence of documentation, support for user constraints, and support for inductive learning.

| Methods | Open Source | Usable | Docs | Community | Constraints | Inductive | AutoFE |
|---|---|---|---|---|---|---|---|
| 53 | 27/53 (50.09%) | 11/27 (40.74%) | 10/27 (37.04%) | 7/27 (25.93%) | 0/27 (0%) | 20/27 (74.07%) | 0/27 (0%) |

**Conclusion.** From our survey, it is apparent that most AutoFE methods have poor usability, and none support user constraints for memory and time budgets. Thus, we deem none of the surveyed methods to be an AutoFE method that practitioners *should use*. We call upon the community to focus more on usability than methodological advancements in future work to unlock the full potential of AutoFE for practitioners.

## 5 Detailed Overview

Table 2 presents the detailed results of our survey across the following criteria. We use "/" whenever a criterion was not applicable, such as when we were not able to inspect the code quality for methods without code available.

**Usable:**  if we were able to run the method, a  means that there were non-solvable issues, which prevented us from using the method.

**Open-Source:**  if the code is open-source,  in case it is not publicly available, and  if there is no official, but an unofficial implementation of the method.

**Code:**  indicates that the code is structured in an understandable manner, following software quality best practices;  indicates that this is not the case.

**Docs:**  signifies the method has documentation and  means that there is no documentation.

**Community:**  indicates that the method has an active community and  indicates the opposite.

**Year:** states the last year the code was updated. If no code is available, we state the publication year of the paper.

**Effort:** provide an impression on how difficult it is to get the method to run, as detailed in the appendix.  means that using the method is straightforward;  indicates that the effort was higher, which means that there was either no simple installation or some issues in the code that had to be resolved to run the method;  signifies that we encountered unsolvable problems while trying to run the method, such as severe bugs in the code or no feasible dependencies (or description thereof).

**Constraints:**  indicates that the user is allowed to set a time and a memory budget for the AutoFE method;  if the user can set any other constraints, such as the number iteration steps; and  if the method does not support setting any constraints.

**Inductiveness:**  if the method is preventing transductive learning and  if the method is always performing transductive learning.

**AutoFE:** our final judgment whether we consider the method to be automatic feature engineering as described in Section 3.

Table 2: Overview of the evaluated criteria for 53 AutoFE methods.

| Method | Usable | Open Source | Code | Docs | Community | Year | Effort | Constraints | Inductiveness | AutoFE |
|---|---|---|---|---|---|---|---|---|---|---|
| autofeat [13] | ✅ | ✅ | ✅ | ✅ | ❌ | 2025 | 🟢 | 🟡 | ✅ | ❌ |
| Autogluon [25] | ✅ | ✅ | ✅ | ✅ | ✅ | 2025 | 🟢 | 🔴 | ✅ | ❌ |
| BioAutoML [26] | ✅ | ✅ | ❌ | ✅ | ❌ | 2023 | 🟡 | 🟡 | ✅ | ❌ |
| Boruta [27] | ✅ | ✅ | ❌ | ❌ | ✅ | 2024 | 🟡 | 🔴 | ✅ | ❌ |
| CFS [28, 29] | ✅ | ⚠️ | ❌ | ❌ | ❌ | 2023 | 🟡 | 🔴 | ✅ | ❌ |
| Featurewiz [30] | ✅ | ✅ | ❌ | ✅ | ✅ | 2025 | 🟡 | 🔴 | ✅ | ❌ |
| H2O [31] | ✅ | ✅ | ❌ | ✅ | ✅ | 2024 | 🟡 | 🔴 | ❌ | ❌ |
| MACFE [32] | ✅ | ✅ | ✅ | ❌ | ❌ | 2022 | 🟡 | 🟡 | ❌ | ❌ |
| MAFESE [33] | ✅ | ✅ | ✅ | ✅ | ❌ | 2025 | 🟢 | 🟡 | ✅ | ❌ |
| MLJAR [34] | ✅ | ✅ | ✅ | ✅ | ✅ | 2025 | 🟡 | 🟡 | ✅ | ❌ |
| OpenFE [12] | ✅ | ✅ | ✅ | ❌ | ❌ | 2024 | 🟢 | 🔴 | ❌ | ❌ |
| AdaFS [35] | ❌ | ✅ | ❌ | ❌ | ❌ | 2023 | / | 🟡 | ✅ | ❌ |
| AutoField [36] | ❌ | ⚠️ | ❌ | ❌ | ❌ | 2022 | / | 🟡 | ✅ | ❌ |
| Autolearn [37] | ❌ | ✅ | ❌ | ❌ | ❌ | 2018 | 🔴 | 🔴 | ❌ | ❌ |
| CAAFE [5] | ❌ | ✅ | ✅ | ❌ | ❌ | 2024 | 🔴 | 🔴 | ✅ | ❌ |
| DIFER [38] | ❌ | ✅ | ✅ | ❌ | ❌ | 2022 | 🔴 | 🟡 | ✅ | ❌ |
| ExploreKit [14] | ❌ | ✅ | ✅ | ❌ | ❌ | 2016 | 🔴 | 🟡 | ✅ | ❌ |
| Featuretools [39] | ❌ | ✅ | ✅ | ✅ | ✅ | 2024 | 🟡 | 🔴 | ❌ | ❌ |
| FETCH [40] | ❌ | ✅ | ❌ | ❌ | ❌ | 2023 | 🔴 | 🟡 | ❌ | ❌ |
| FGCNN [41] | ❌ | ✅ | ❌ | ❌ | ❌ | 2021 | 🔴 | 🟡 | ❌ | ❌ |
| LAMA [42] | ❌ | ✅ | ✅ | ✅ | ✅ | 2025 | 🔴 | 🔴 | ✅ | ❌ |
| LPFS [43] | ❌ | ⚠️ | ❌ | ❌ | ❌ | 2022 | / | 🟡 | ✅ | ❌ |
| MFE [44] | ❌ | ✅ | ✅ | ✅ | ❌ | 2022 | 🔴 | 🔴 | ✅ | ❌ |
| NewApproachFS [45] | ❌ | ✅ | ❌ | ❌ | ❌ | 2022 | 🔴 | 🔴 | ✅ | ❌ |
| NFS [46] | ❌ | ⚠️ | ✅ | ❌ | ❌ | 2019 | 🟡 | 🟡 | ✅ | ❌ |
| OptFS [47] | ❌ | ✅ | ✅ | ❌ | ❌ | 2022 | 🔴 | 🟡 | ✅ | ❌ |
| SimplerIsBetter [48] | ❌ | ✅ | ✅ | ❌ | ❌ | 2021 | 🔴 | 🔴 | ✅ | ❌ |
| AEFE [49] | / | ❌ | / | / | / | 2021 | / | / | / | / |
| AFER [50] | / | ❌ | / | / | / | 2023 | / | / | / | / |
| ALSFS [51] | / | ❌ | / | / | / | 2017 | / | / | / | / |
| ANOFS [52] | / | ❌ | / | / | / | 2015 | / | / | / | / |
| ARLFS [53] | / | ❌ | / | / | / | 2009 | / | / | / | / |
| AutoCross [54] | / | ❌ | / | / | / | 2019 | / | / | / | / |
| AutoFS [55] | / | ❌ | / | / | / | 2020 | / | / | / | / |
| AutoGFS [56] | / | ❌ | / | / | / | 2021 | / | / | / | / |
| AutoModeling [57] | / | ❌ | / | / | / | 2018 | / | / | / | / |
| Brainwash [58] | / | ❌ | / | / | / | 2013 | / | / | / | / |
| FSTD [59] | / | ❌ | / | / | / | 2019 | / | / | / | / |
| FuzzyFS [60] | / | ❌ | / | / | / | 2019 | / | / | / | / |
| GLIDER [61] | / | ❌ | / | / | / | 2020 | / | / | / | / |
| GPFS [62] | / | ❌ | / | / | / | 2018 | / | / | / | / |
| LassoFS [63] | / | ❌ | / | / | / | 2017 | / | / | / | / |
| LFS [64] | / | ❌ | / | / | / | 2018 | / | / | / | / |
| MARLFS [15] | / | ❌ | / | / | / | 2022 | / | / | / | / |
| MEFS [65] | / | ❌ | / | / | / | 2022 | / | / | / | / |
| MeLiF [66] | / | ❌ | / | / | / | 2016 | / | / | / | / |
| MFS [67] | / | ❌ | / | / | / | 2014 | / | / | / | / |
| RegularizedFS [68] | / | ❌ | / | / | / | 2017 | / | / | / | / |
| Relief [69] | / | ❌ | / | / | / | 2018 | / | / | / | / |
| RLFS [16] | / | ❌ | / | / | / | 2012 | / | / | / | / |
| ReinSel [70] | / | ❌ | / | / | / | 2010 | / | / | / | / |
| SARLFS [17] | / | ❌ | / | / | / | 2020 | / | / | / | / |
| SVMFS [71] | / | ❌ | / | / | / | 2009 | / | / | / | / |

**Acknowledgements.** L.P. acknowledges funding by the Deutsche Forschungsgemeinschaft (DFG, German Research Foundation) under SFB 1597 (SmallData), grant number 499552394. F.H. acknowledges the financial support of the Hector Foundation. M.J. acknowledges the funding by The Carl Zeiss Foundation through the research network "Responsive and Scalable Learning for Robots Assisting Humans" (ReScaLe) of the University of Freiburg. The authors gratefully acknowledge the scientific support and HPC resources provided by the Erlangen National High Performance Computing Center (NHR@FAU) of the Friedrich-Alexander-Universität Erlangen-Nürnberg (FAU).

## A Method Investigation - Methodology

In the following, the method investigation is explained in detail for each method. We provide a short method description, present our results of the analysis of the code structure, the documentation, the communty, the user constraints and the transductiveness and finally talk about the pracitcal usage of the methods.

### A.1 AdaFS [35]

**A.1.1 Method Description.** The main idea of AdaFS is to find a subset of features with an adaptable size by having a controller network selecting the best features out of the feature space. This is achieved by an adapted controller scoring features for each recommendation task.

**A.1.2 Code Structure, Documentation, Community, Constraints & Transductiveness.** The method has official open-source code from 2023 (AdaFS - GitHub), which is not well structured considering the usual structure of source code including a source, data, external scripts and documentation folder. A data folder is provided containing one training dataset. There exists also an unofficial version (AdaFS - Unofficial GitHub), where the structure is also not given and no data is provided. No community is active in contributing or using the software and no documentation is provided to the user. Regarding the constraints, the user can limit the number of epochs and the batch size, when using AdaFS. The code prevents transductive learning.

**A.1.3 Installation & Utilization.** AdaFS is not installable via PyPI. The problem when using the method in this context is, that AdaFS is not fitting the use case: There is a soft and a hard version, while the soft version is doing a backpropagation weighting the features, the hard version is giving the features with the lowest feature values a zero embedding. In both versions, the feature values are assigned based on the importance of the model output, which is not available for feature engineering in this context. For this reason, the method is not usable in our context.

**A.1.4 Automation.** The method is not an AutoFE method, since it is not usable with minimal effort and without immense expertise and the required resource usage constraints are not fulfilled.

### A.2 AEFE [49]

**A.2.1 Method Description.** AEFE is a framework with the purpose of representing categorical features using modules like custom paradigm feature construction and multiple feature selection. It generates interpretable combinations of features and determines the feature importance.

**A.2.2 Code Structure, Documentation, Community, Constraints & Transductiveness.** There is no open-source code, which means, that the code structure can not be evaluated, as well as the user constraints and the transductiveness. We can also not check if there is a documentation or a community.

**A.2.3 Installation & Utilization.** This method does not have any publicly available code, which is why it is not usable.

**A.2.4 Automation.** The method is not an AutoFE method, since it is not usable with minimal effort and without immense expertise. The required resource usage constraints can not be evaluated.

### A.3 AFER [50]

**A.3.1 Method Description.** The "Automated Feature Engineering for Robotic Prediction on Intelligent Automation"-approach uses feature engineering for intelligent automation in the context of robot process automation by using an elastic net regression based algorithm.

**A.3.2 Code Structure, Documentation, Community, Constraints & Transductiveness.** The code of this method is not publicly available, which makes it not possible to assess the code structure, the documentation, the community, the user constraints and the transductiveness.

**A.3.3 Installation & Utilization.** Getting a working implementation is not feasible, as the method is not open-source. For this reason, it is not usable.

**A.3.4 Automation.** The method is not an AutoFE method, since it is not usable with minimal effort and without immense expertise. The required resource usage constraints can not be evaluated.

## A.4 ALSFS [51]

**A.4.1 Method Description.** ALSFS is a method for "Improving Activity Recognition Accuracy in Ambient-Assisted Living Systems by Automated Feature Engineering". The authors propose a feature extraction and selection procedure especially focusing on time-variant features.

**A.4.2 Code Structure, Documentation, Community, Constraints & Transductiveness.** The lack of public access to the code makes it impossible to evaluate its structure, documentation, community, user constraints, or transductiveness.

**A.4.3 Installation & Utilization.** Since the method is not open-source, it is not further investigated and not usable.

**A.4.4 Automation.** The method is not an AutoFE method, since it is not usable with minimal effort and without immense expertise. The required resource usage constraints can not be evaluated.

## A.5 ANOFS [52]

**A.5.1 Method Description.** The method ANOFS is focusing on real-world applicable feature selection using online learning and claims to improve the selection by introducing a negotiation theory-based approach.

**A.5.2 Code Structure, Documentation, Community, Constraints & Transductiveness.** Without open-source code, evaluating its structure, documentation, community, user constraints, or transductiveness is not possible.

**A.5.3 Installation & Utilization.** Getting a working implementation is not feasible as the method is not open-source, which is why it is not usable.

**A.5.4 Automation.** The method is not an AutoFE method, since it is not usable with minimal effort and without immense expertise. The required resource usage constraints can not be evaluated.

## A.6 ARLFS [53]

**A.6.1 Method Description.** ARLFS aims to loosen the restrictions of supervised learning by introducing a feature selection approach for reinforcement learning. A Bayesian network of a factored Markov Decision Process is learned as a model by the agent, who selects features from the network and extracts a policy of the feature set, which allows the agent to deduce minimal feature sets.

**A.6.2 Code Structure, Documentation, Community, Constraints & Transductiveness.** The absence of public access to the code prevents the evaluation of its structure, documentation, community, user constraints, or transductiveness.

**A.6.3 Installation & Utilization.** This method is not usable in this context, as there is no code publicly available.

A.6.4 **Automation.** The method is not an AutoFE method, since it is not usable with minimal effort and without immense expertise. The required resource usage constraints can not be evaluated.

### A.7 AutoCross [54]

A.7.1 **Method Description.** The AutoCross-method claims to capture interactions of categorical features by generating high-order cross features with beam search. The authors attempt to improve this method by using successive mini-batch gradient descent and multi-granularity discretization.

A.7.2 **Code Structure, Documentation, Community, Constraints & Transductiveness.** The unavailability of public code hinders the evaluation of its structure, documentation, community, user constraints, and transductiveness.

A.7.3 **Installation & Utilization.** This method does not have any publicly available code, which is the reason for not further investigating and deciding it is not usable.

A.7.4 **Automation.** The method is not an AutoFE method, since it is not usable with minimal effort and without immense expertise. The required resource usage constraints can not be evaluated.

### A.8 autofeat [13]

A.8.1 **Method Description.** autofeat is a library for automated feature generation and feature selection providing a linear regression model for understandable and easily trainable AutoFE.

A.8.2 **Code Structure, Documentation, Community, Constraints & Transductiveness.** The library last updated 2025 is publicly available (GitHub) and the code is structured following the given scheme. There are some datasets included in usage examples and the authors do provide a basic documentation with the code (autofeat - Documentation). Besides the 5 authors of the repository, there is no community. As constraints, the user can set the number of steps for feature generation and feature selection as well as the operators for feature generation. autofeat is not transductive.

A.8.3 **Installation & Utilization.** The code of the method is usable in python by installing it via the official third-party software repository PyPI. The usage of the library can be conducted straightforward by creating an instance of the "AutoFeatRegressor" or "AutoFeatClassifier" - depending on the task, which can be fit on the training data and be used to transform training and test data. Including autofeat does not cause any issues and the method is for that usable.

A.8.4 **Automation.** The method is not an AutoFE method, since the required resource usage constraints are not fulfilled.

### A.9 AutoField [36]

A.9.1 **Method Description.** For the feature selection in deep learning-based recommender systems, AutoField suggests a framework composed of a differentiable controller network for selecting feature fields.

A.9.2 **Code Structure, Documentation, Community, Constraints & Transductiveness.** AutoField does not have an official GitHub, but there is an unofficial open-source version, which can be compared to the method in the paper (AutoField - Unofficial GitHub). The code does not follow the expected structure and there is no data example, no community and no documentation. In the code of the unofficial GitHub repository, the user can set a percentage for the dropout and the batch size as user constraints. The unofficial AutoField implementation does not prevent transductive learning.

**A.9.3 Installation & Utilization.** In the code, there is a feature selection "Module", a "Deep Recommendation Model" and a "controller" to perform feature selection based on the model output (AutoField-Unofficial GitHub). The problem is, that in this setting, the model output is not accessible for the feature engineering, which is the reason for deciding, that this is method is not usable in this context.

**A.9.4 Automation.** The method is not an AutoFE method, since it is not usable with minimal effort and without immense expertise and the required resource usage constraints are not fulfilled.

## A.10 AutoFS [55]

**A.10.1 Method Description.** AutoFS is a feature space navigation method which uses the idea of interactive reinforced feature selection. In this scenario, a self-exploring but also externally trained agent performs feature selection.

**A.10.2 Code Structure, Documentation, Community, Constraints & Transductiveness.** Since the code is not open-source, we cannot assess its structure, user constraints, or transductiveness. Additionally, we are unable to verify the availability of a documentation or public data.

**A.10.3 Installation & Utilization.** Since the method is not open-source, it is not further investigated and not usable.

**A.10.4 Automation.** The method is not an AutoFE method, since it is not usable with minimal effort and without immense expertise. The required resource usage constraints can not be evaluated.

## A.11 AutoGFS [56]

**A.11.1 Method Description.** AutoGFS aims to trade-off single-agent and multi-agent reinforced feature selection by proposing a group-based interactive reinforced feature selection framework. This approach divides features into groups, each belonging to an agent, which can decide which feature to include and which features to exclude from the group. Additionally, there is a training agent performing intra- and inter-group selection.

**A.11.2 Code Structure, Documentation, Community, Constraints & Transductiveness.** There is no open-source code, which means, that the code structure can not be evaluated, as well as the user constraints and the transductiveness. We can also not check if there is a documentation or a community.

**A.11.3 Installation & Utilization.** This method does not have any publicly available code, which is the reason for not further investigating it and deciding it is not usable.

**A.11.4 Automation.** The method is not an AutoFE method, since it is not usable with minimal effort and without immense expertise. The required resource usage constraints can not be evaluated.

## A.12 Autogluon [25]

**A.12.1 Method Description.** The AutoML framework Autogluon ensembles multiple models stacked to multiple layers. Within this architecture, there exists a part intended to do AutoFE, which is used in this setting as an AutoFE method. The feature engineering procedure in Autogluon is not mentioned in the paper, the authors only describe the model-agnostic and model-specific data preprocessing, which consists of factorization and replacing missing values.

**A.12.2 Code Structure, Documentation, Community, Constraints & Transductiveness.** The open-source repository of the method (Autogluon - Official GitHub) is well structured, provides examples and has a active community with 121 contributors and 777 users. The code is last updated in 2025. The code comes with an extensive documentation (Autogluon - Documentation). There are no possibilities for the user to adapt the AutoFE method. The feature engineering is not transductive.

**A.12.3 Installation & Utilization.** The framework Autogluon comes with a "AutoMLPipelineFeatureGenerator" and a "FeatureSelector", which can be extracted form the AutoML framework. Autogluon can be installed via PyPI and it is easily usable by instancing the aforementioned model-agnostic feature generator, fitting it on the training data and transforming training and test data afterwards. The feature selector is model-specific and can not be used in this context, because the model output is needed but not available in this context. The feature generator is working without any problems and is for that usable.

**A.12.4 Automation.** The method is not an AutoFE method, since the required resource usage constraints are not fulfilled.

## A.13 Autolearn [37]

**A.13.1 Method Description.** The regression-based feature learning algorithm Autolearn aims to learn the pairwise feature association and selects the stable and performance-improving relationships.

**A.13.2 Code Structure, Documentation, Community, Constraints & Transductiveness.** The code of the paper form 2018 is publicly available on GitHub (Autolearn - GitHub). The code is not structured, there only exists one file, but the author provides an example dataset. There is no documentation and no community. A potential user can not make any constraints to the method and Autolearn is transductive.

**A.13.3 Installation & Utilization.** Autolearn is not installable via PyPI. When integrating the method in this context, there occur several issues:
```
1. ImportError: cannot import name 'perceptron' from 'sklearn.linear_model'
```
This import error can be fixed by correcting the spelling of perceptron, with a capital P, after that it can be imported.
```
2. ImportError: cannot import name 'RandomizedLasso' from 'sklearn.linear_model'
```
This issue can not be fixed, Autolearn is working with an old version of sklearn, that could not be installed because there are "no wheels for building scikit-learn 0.15". For that reason, the method is not usable.

**A.13.4 Automation.** The method is not an AutoFE method, since it is not usable with minimal effort and without immense expertise and the required resource usage constraints are not fulfilled.

## A.14 AutoModeling [57]

**A.14.1 Method Description.** The method AutoModeling aims on selecting the ensemble model consisting of a selected feature subset and a selected pretrained classifier by using a relax-greedy search.

**A.14.2 Code Structure, Documentation, Community, Constraints & Transductiveness.** The absence of public access to the code prevents any evaluation of its structure, documentation, community, user constraints, or transductiveness.

**A.14.3 Installation & Utilization.** Getting a working implementation is not feasible, as the method is not open-source it is not usable.

**A.14.4 Automation.** The method is not an AutoFE method, since it is not usable with minimal effort and without immense expertise. The required resource usage constraints can not be evaluated.

## A.15 BioAutoML [26]

### A.15.1 Method Description
BioAutoML is a AutoML framework for biological sequence data consisting of an AutoFE component including the extraction of numerical features and automatic feature selection and of meta-learning including machine learning algorithm recommendation and tuning of the hyperparameters of the selected algorithm. In this context, we are not interested in the AutoML part, which is why only the AutoFE is extracted from the code.

### A.15.2 Code Structure, Documentation, Community, Constraints & Transductiveness
The authors provide a documentation (BioAutoML - Documentation) together with the publicly available code (BioAutoML - GitHub). Even if there are some examples, the code is not structured and there is no community. The user can set the number of estimations in order to limit the AutoFE algorithm. Transductive Learning is prevented.

### A.15.3 Installation & Utilization
The BioAutoML software from 2023 is not installable via PyPI, which forces the user to copy the code from the GitHub repository. In the file "BioAutoML-feature.py" the "feature_engineering"-function can be found, which is applicable for our case. While trying to use the function, there are four problems: At first, the main-method in the "BioAutoML-feature.py" is not executable since the parser arguments can not be initialized properly and there is no documentation provided to help the user. By utilizing only the "feature_engineering" function, we can solve this issue.
```
2.  IndexError:  positional indexers are out-of-bounds
```
Secondly, when inputting the training data we are running into an "index out-of-bounds" error. The method seems to need the full dataset for the cross-validation conducted while minimizing the function over the hyperparameter space.
```
3.  hyperopt.exceptions.AllTrialsFailed
```
The next problem is that all "hyperopt Trials" fail, when the "fasta_label_train" variable is set to a value bigger than two. The variable determines the model, if the value is bigger than 2, the used model is a "AdaBoostClassifier", otherwise the "CatBoostClassifier" is used. With using the "AdaBoostClassifier", the execution fails. When using the "CatBoostClassifier", the function is running into a problem with categorical values.
```
4.  CatBoostError:  Invalid label type=<class
'pandas.core.arrays.categorical.Categorical'>
```
The error with categorical values can be fixed by factorizing all categorical columns of the dataset, which turns each category into a number representing such category. After solving these issues, the method works and is usable.

### A.15.4 Automation
The method is not an AutoFE method, since it is not usable with minimal effort and without immense expertise and the required resource usage constraints are not fulfilled.

## A.16 Boruta [27]

### A.16.1 Method Description
The approach Boruta is an AutoFE algorithm removing all features that are less relevant as randomly generated features with a Random Forest classifier. As stated in the blogpost, the former R code was transferred to python, which increased the speed of the method and made it usable in this case (Boruta - Blogpost).

### A.16.2 Code Structure, Documentation, Community, Constraints & Transductiveness
The 2024 updated code for Boruta is publicly available (GitHub). It is not structured well, but there are some examples provided. There is no documentation, but the repository has a big community of more than 2000 users and 20 contributors. The user can constraint the number of estimators, the maximal number of iterations and a early stopping criterion to restrict the AutoFE method. The method prevents transductive learning.

**A.16.3 Installation & Utilization.** The software can be installed via PyPI . An instance of the "BorutaPy" feature selector can be fitted on the training data and transform the training and test data likewise. While integrating this procedure in the context, two errors occurred:
`1. AttributeError: module 'numpy' has no attribute 'int'`
The code has data type specification like "np.int", which are not working and have to be updated to "np.int64", which makes it necessary to fork the code and locally apply the changes.
`2. ValueError: could not convert string to float`
The second problem that needs to be solved is the treatment of categorical and non-numerical "NaN" values, which can be solved by factorization and imputing. With these fixes, the method is working and is usable.

**A.16.4 Automation.** The method is not an AutoFE method, since it is not usable with minimal effort and without immense expertise and the required resource usage constraints are not fulfilled.

## A.17 Brainwash [58]

**A.17.1 Method Description.** The AutoFE method Brainwash follows the Explore-Extract-Evaluate principle. In the Explore part it collects statistical information on the datasets, in the Extract phase, Brainwash selects features by using feature induction and in the evaluation, the method checks the performance of the feature engineered dataset with the model.

**A.17.2 Code Structure, Documentation, Community, Constraints & Transductiveness.** The unavailability of public code hinders the evaluation of its structure, documentation, community, user constraints, and transductiveness.

**A.17.3 Installation & Utilization.** Since the method is not open-source, it is not further investigated and not usable.

**A.17.4 Automation.** The method is not an AutoFE method, since it is not usable with minimal effort and without immense expertise. The required resource usage constraints can not be evaluated.

## A.18 CAAFE [5]

**A.18.1 Method Description.** Based on the description of tabular datasets, the AutoFE method CAAFE incorporates domain knowledge in AutoML by using a large language model (LLM) in order to generate new semantically meaningful features.

**A.18.2 Code Structure, Documentation, Community, Constraints & Transductiveness.** CAAFE is an open-source project (GitHub) last updated 2024 with a well-maintained and structured code including data examples. There is no community contributing to the project. The authors do not provide a documentation. The user can set the number of repetitions in order to limit the runtime of the method. CAAFE prevents transductive learning.

**A.18.3 Installation & Utilization.** CAAFE is an open-source project and can be installed by using PyPI . Within the installation process, PyPI uninstalls the previously installed version of OpenML ([[72]] 0.14.2 and installed the version 0.12.0 instead, which does not include the "get_task"-function, that is used in this project to retrieve the datasets. Manually reinstalling the Openml 0.14.2 version fixed the issue and did not interfere with CAAFE. The method can be used by instancing the "CAAFEClassifier" and fit it on the training data. There also exists a "generate_features"-function that can be used for generating code, prompt and messages using a LLM. The problem with this method is, that it can not be run locally and the overhead is extremely high as it uses a LLM. The usage of CAAFE is too expensive and for that reason the tool is not usable.

**A.18.4 Automation.** The method is not an AutoFE method, since it is not usable with minimal effort and without immense expertise and the required resource usage constraints are not fulfilled.

## A.19 CFS [28, 29]

**A.19.1 Method Description.** CFS is a correlation based approach to feature selection based on the assumption, that good features are highly correlated with the target but not correlated with each other. The publicly available code combines the idea of this paper with the CFS-method introduced by Yu in 2003, who proposes the identification of relevant features and of redundancy amongst them by using a fast filter method.

**A.19.2 Code Structure, Documentation, Community, Constraints & Transductiveness.** The code from 2023 for the joint Correlation-based Feature Selection method is publicly available (CFS - Unofficial GitHub) The code is not structured well, there is no public data and no community supporting the project as well as no documentation guiding the user. There are no constraints possible to the AutoFE method when using the "cfs" function, but a threshold can be set for the "fcbf" function. CFS is not transductive.

**A.19.3 Installation & Utilization.** The code is not installable via PyPI. When forking the project and using the code, which means instancing the "MUFS" class and using its "cfs"- or "fcbf"-function, there are two problems:
```
1.   TypeError:   axis argument not supported
```
At first, there is a problem with the Dataset, the pandas "DataFrame" used as dataset data type in this project can not be used with this method. To solve this issue, it needs to be converted into a "numpy.ndarray".
```
2.   ValueError:   Expected 2D array, got 1D array instead
```
Secondly, the target "numpy.ndarray" is posing a problem, because the shape are not as it is expected by the method (e.g. (621, 1) instead of (621,)). To resolve this issue, the target array needs to be flattened. With these changes, the method works and is usable.

**A.19.4 Automation.** The method is not an AutoFE method, since it is not usable with minimal effort and without immense expertise and the required resource usage constraints are not fulfilled.

## A.20 DIFER [38]

**A.20.1 Method Description.** The differentiable AutoFE method DIFER iteratively maps features into the continuous vector space using a encoder, the predictor improves the embedding and the decoder retrieves the features.

**A.20.2 Code Structure, Documentation, Community, Constraints & Transductiveness.** The code for DIFER is available on GitHub (DIFER - GitHub) and was last updated 2022. The code is organized; however, it does not conform to the specified format. Nonetheless, data is available for use. There is no documentation provided and DIFER does not have a community. The user can restrict the number of estimators and the number of total iterations in order to limit the AutoFE method. Additionally, a time budget can be set. DIFER prevents transductive learning.

**A.20.3 Installation & Utilization.** The code is not installable via PyPI. In order to use the method in our setting, the user needs to convert the given pandas "DataFrame" to a "MixDataset", prepare the "SklearnEnv" environment, initiallize "Tokenizer", "NFOController" and "FeatureGenerator" and transform the new features from the "FeatureGenerator" with the "Tokenizer". In the end, the best features are selected using the "NFOController". When including this method in our context, there are several issues:
```
1.   Error in construct
2,2,division,2,-,2,reciprocal,reciprocal,+,2,log,- 2
```
The first error occurring is the error in the feature construction, which also produces a Key Error with a different key in every execution. It is not possible to debug the working code using 3 of the datasets provided by the authors as the error is occurring there, too. As the error message is likely to arise in the step of feature generation, this procedure needs to be investigated. In the file "feat_tree.py" exists the function "generate(self, env)". This function has one condition, if the variable "self.arity" equals 0, the feature is converted to an "Integer" and the dataset is converted to a "numpy.ndarray", if it is not equal to 0 new features are generated by the following code in line 63:

```
children_feats = [each.generate(env) for each in self.children if each is not None]
```

The IDE used in this context (PyCharm) is not able to find a "generate(env)" function, that exists for "self.children". In the debugging procedure, we are also not able to find the corresponding code. The value of "self.arity" is the indicator for what is happening, if "self.arity" is not equal to zero, feature construction is executed; however, if "self.arity" is zero, the feature construction process fails. Since in this context, we constrain methods to be practicably usable within a reasonable amount of time and we were not able to execute the method with the datasets provided by the method itself. DIFER is not usable.

A.20.4 **Automation**. The method is not an AutoFE method, since it is not usable with minimal effort and without immense expertise and the required resource usage constraints are not fulfilled.

## A.21 ExploreKit [14]

A.21.1 **Method Description**. By generating features based on combination of original features and selecting them following a machine learning-based prediction of their usefulness, the framework ExploreKit performs AutoFE.

A.21.2 **Code Structure, Documentation, Community, Constraints & Transductiveness**. The code for ExploreKit is publicly available on GitHub (ExploreKit - GitHub). The code is structured in an understandable manner, but there is no example data provided. The authors do not provide a documentation and there is no community. The user can restrict the AutoFE method by setting the number of maximal iterations. ExploreKit prevents transductive learning.

A.21.3 **Installation & Utilization**. The code is written in Java, which means it is not usable in the python environment of this context, but there exists an unofficial version of ExploreKit in python, which appears to be a direct transformation from the Java code to python and can for that be considered here (ExploreKit - Unofficial GitHub). The unofficial code is last updated in 2016 and not installable via PyPI, the user needs to copy the files. For using the method, the "FilterWrapperHeuristicSearch" needs to be instanced and the corresponding run method can be used in order to get the dataset and the candidate features. There are several issues when including ExploreKit in the context:

```
1. TypeError: module() takes at most 2 arguments (3 given)
```

The type error is occuring when calling "FilterWrapperHeuristicSearch(15)" without any changes to the code. There is no issue on GitHub regarding this problem, but it can be fixed by changing the imported libraries in the corresponding file.

```
2. FileNotFoundError: [Errno 2] No such file or directory
```

This file not found error can be resolved by changing "os.mkdir" to "os.makedirs" in line 192 in the function "createDatasetMetaFeaturesInstances" in the file "Evaluation/MLAttributeManager.py".

```
3. ValueError: setting an array element with a sequence. The
requested array has an inhomogeneous shape after 1 dimensions.
The detected shape was (112,) + inhomogeneous part
```

In order to solve this value error, the user needs to change the line 671 in the function "_generateValuesMatrix" in the file "MLArributeManager.py" "df = pd.DataFrame(np.asarray(attributesMatrix), columns=columns)" to "df = pd.DataFrame(np.asarray(attributesMatrix, dtype=object)) #, columns=columns)".

4. `arff.Bad Object Error: Invalid attribute declaration "(0, ['768'])"`

The following error is a bad object error of the "arff" library and it is occurring with all our datasets and all datasets provided together with the code by the authors of the library. In the file "MLAttributeManager.py" there exists a function called "generateTrainingSetDatasetAttributes", which is used for generating "datasetAttributes". The idea is to do so and replicate the dataset, create discretized features and add them to the dataset. We obtain non-unary operator assignments taking advantage of the discretized features, but the feature generation fails because a "NoneType" object has no attribute "getEvaluationStats". One other idea would be to skip the whole procedure of generating the model and get it from somewhere else. As the Java code is doing a similar thing, the model generated there could be similar to the one we are not able to create using the python code. When trying to execute the Java code, there are problems with imports that can not be fixed, as it can be seen in the Issues subsection of the corresponding GitHub repository. The model can not be retrieved directly from the repository or any other source. There are no answers and no documentation is provided with the code to solve the issue, which is why ExploreKit is not usable.

**A.21.4 Automation**. The method is not an AutoFE method, since it is not usable with minimal effort and without immense expertise and the required resource usage constraints are not fulfilled.

## A.22 Featuretools [39]

**A.22.1 Method Description**. Featuretools is an automatic feature generation method for relational datasets. The method creates features by ensuing relationships in the data and utilize mathematical functions as transformations.

**A.22.2 Code Structure, Documentation, Community, Constraints & Transductiveness**. The code from 2024 of Featuretools is publicly available (Featuretools - GitHub), it is understandably structured, but there is no public data. There exists an extensive documentation for the method (Featuretools - Documentation) and a large community of more than 70 contributors and almost 2000 users. The user can not constrain the execution of the AutoFE method. The method does not prevent transductive learning.

**A.22.3 Installation & Utilization**. Featuretools can be installed via PyPI. At first, we need to generate a "EnitjtySet" from the pandas "DataFrame" that we have as a input for the AutoFE method by instancing the "EntitySet" and adding the training data to it. Then the "dfs"-function can be used that calculates a feature matrix and features. The fist problem is that the method is running without errors but not changing the dataset. One can think of different aspects of the method in order to understand why it is not doing what we expect it to do: Firstly, the operators could be wrong, since they are not clear (e.g. "day"), but some of them are reasonable (e.g. ("haversine") and there is a default list of operations that is always used, so that should not be the issue. Another idea are wrong primitives, which are aggregation operations applied on features like "sum", but they are also working. It turns out that the relationships between datasets seem to be not necessary in our case, but Featuretools needs relationships in order to work. In addition to that, the user can also set a target. Adding the relationships is not trivial, since there is no logical need for relationships in our dataset and there is no documentation on how to create relationships, there are four combinatorial possibilities: At first, we could combine dataset and target by a relationship leaving the target set to the y component of our training dataset, but in this case, the output is only one feature counting the appearances of different values in the target. Secondly, we try to combine the train and the test dataset letting the target being the y part of the training dataset, which also did not lead to any useful features. Also combining two features of the same dataset does not change the outcome, but combining two features of the same training dataset and setting the training dataset as the target is generating a lot of features. In the end, there is one parameter causing memory issues, which is the max_depth variable.
```
1. RecursionError: maximum recursion depth exceeded while
pickling an object
```
If the max_depth is set to 10, the recursion error occurs. By setting it to 2, which is also the default, there are the same memory issues, no recursion error. Setting it to the minimal value, which is 1, we are running in the same problems. This issue is not fixable within the scope of this paper, which is the reason for deciding that this method is not usable within this context.

**A.22.4 Automation.** The method is not an AutoFE method, since it is not usable with minimal effort and without immense expertise and the required resource usage constraints are not fulfilled.

### A.23 Featurewiz [30]

**A.23.1 Method Description.** Featurewiz is a content-based image retrieval framework fusing several feature descriptors in order to get a better retrieval performance. The method extracts key points separately from the images, which are mapped to a single keypoint feature vector. In the end, the best features are selected for a bag-of-visual-words learning model.

**A.23.2 Code Structure, Documentation, Community, Constraints & Transductiveness.** The open-source code (Featurewiz - GitHub) maintained in 2025 is not structured, but the data is publicly available. There is a documentation provided (Featurewiz - Documentation), and there is a community of 16 contributors. The user can not make any constraints to the execution of the method. Featurewiz prevents transductive learning.

**A.23.3 Installation & Utilization.** Even if Featurewiz is initially used for images, the code can be applied for tabular data. Featurewiz is installable via PyPI, but the installation process is causing problems: Featurewiz requires different versions of the "scikit-learn" library than used in this context, which breaks the cross-validation evaluation part of the AutoML-Toolkit Pipeline after the installation. In order to fix this issue, it is necessary to manually re-install the latest "scikit-learn" version used before in this context and fix the errors occuring within the Featurewiz method by manually updating the outdated usage of the "scikit-learn" library. The application of the AutoFE method can be achieved by instantiating the "FeatureWiz" class, fitting it to the training data, and subsequently transforming both the training and test datasets accordingly. There is one error when including the code in this context:
```
1. ImportError: cannot import name 'if_delegate_has_method' from
'sklearn.utils.metaestimators'
```
The error occuring here is an import error for the "if_delegate_has_method" attribute, which can be replaced with the current attribute "available_if" covering the same functionality. With this changes, the method is properly running and for that usable.

**A.23.4 Automation.** The method is not an AutoFE method, since it is not usable with minimal effort and without immense expertise and the required resource usage constraints are not fulfilled.

### A.24 FETCH [40]

**A.24.1 Method Description.** The AutoFE framework "Feature Set Data-Driven Search" (FETCH) is a Markov Decision Process using the given dataset as state for a pretrained policy network.

**A.24.2 Code Structure, Documentation, Community, Constraints & Transductiveness.** The code, which is publicly available on GitHub (FETCH - GitHub) and was last updated 2021 is not well structured. There is data publicly available. The authors do not provide a documentation and there is no community supporting the project. The user can restrict the number of epochs and episodes, the batch size and the number of steps. FETCH does not prevent transductive learning.

- **A.24.3 Installation & Utilization.** The code can not be installed with PyPI. The user is required to download the code from GitHub. An instance of "AutoFE" can be fitted for searching the best AutoFE strategy of attention. This instance can then be used to transform the training and test data by providing the two variables "actions_c" and "actions_d". While trying to include this code in our context, there are some issues:
  ```
  1.   KeyError:  'label'
  ```
  This error seems related to the variables "actions_c" and "actions_d" passed to the transform function. There is no documentation for the meaning of the variable, in the paper it is stated, that "actions_c" is a feature engineering plan consisting of m feature transformation actions. A more detailed investigation reveals, that the problem is the initialization of the pipeline, when using our data. With the data supplied by the authors, the pipeline operates without issues; however, the two previously mentioned variables present a challenge. Setting the values using the corresponding setter methods is not working here, referencing all worker.actions does not work as "workers_top5" is empty. This problem can not be solved, since there is no documentation for understanding how to configure the two actions variables and for this reason, FETCH is not usable.

## A.25 FGCNN [41]

- **A.25.1 Method Description.** FGCNN proposes AutoFE based on convolutional neural networks following the expand and reduce approach with two components, the generator and the deep classifier. The feature generator is a Convolutional Neural Network searching for local patterns and combine them to new features. The deep classifier learns interactions from the augmented feature space.

- **A.25.2 Code Structure, Documentation, Community, Constraints & Transductiveness.** The code is publicly available on GitHub (FGCNN - GitHub). It is not structured and there is no public data. There exists no documentation for the code of FGCNN and the project does not have a community. The user can adjust the configuration of the Convolutional Neural Network and by that restrict the method. The method does not prevent transductive learning.

- **A.25.3 Installation & Utilization.** The FGCNN method is not installable via PyPI. In order to use it, one has to instantiate the "FGCNN" class and pass it to a "TabularLearner" instance, which can be fitted on the training data and can transform the training and test data. There are three problems occurring while trying to make this code usable in this context.
  ```
  1.   The code is written for one dataset
  ```
  In the code, the authors hard-coded the feature names, which is unusual, but easily replaceable by a more generic solution where the feature names are extracted from the pandas "DataFrame" with the columns attribute.
  ```
  2.   There is a self-implemented "DataLoader"
  ```
  The DataLoader the authors implemented by themselves is not working with the pandas "DataFrames" that are used in this context, which makes it necessary to reimplement this parts.
  ```
  3.   The dataset used in the code is not publicly available
  ```
  Since the dataset used in the code is not open-source, it can not be used for debugging the code in order to understand how it is supposed to work. This makes it much harder to understand how the "DataLoader" is working and how the code needs to be adapted to fit our case. With these two problems, the code is not usable without a reimplementation, which is not feasible in the scope of this paper.

- **A.25.4 Automation.** The method is not an AutoFE method, since it is not usable with minimal effort and without immense expertise. The required resource usage constraints can not be evaluated.

## A.26 FSTD [59]

- **A.26.1 Method Description.** Considering feature selection as a reinforcement learning problem is the approach of FSTD. In order to go through the search space, which is considered to be a Markov

Decision Process, temporal difference is applied and a subset of features is selected by using optimal graph search.

**A.26.2 Code Structure, Documentation, Community, Constraints & Transductiveness.** The code of this method is not publicly available, which makes it not possible to evaluate the code structure, the documentation, the community, the user constraints and the transductiveness.

**A.26.3 Installation & Utilization.** Since the method is not open-source, it is not further investigated and not usable.

**A.26.4 Automation.** The method is not an AutoFE method, since it is not usable with minimal effort and without immense expertise. The required resource usage constraints can not be evaluated.

## A.27 FuzzyFS [60]

**A.27.1 Method Description.** FuzzyFS is a feature selection method based on the fuzzy set theory in order to tackle the lack of flexibility in statistical feature selection methods and move towards human-like fuzzy decisions.

**A.27.2 Code Structure, Documentation, Community, Constraints & Transductiveness.** The absence of public access to the code prevents any evaluation of its structure, documentation, community, user constraints, or transductiveness.

**A.27.3 Installation & Utilization.** Getting a working implementation is not feasible, as the method is not open-source. For that reason, it is not usable.

**A.27.4 Automation.** The method is not an AutoFE method, since it is not usable with minimal effort and without immense expertise. The required resource usage constraints can not be evaluated.

## A.28 GLIDER [61]

**A.28.1 Method Description.** The AutoFE method GLIDER aims on generating interpretable features that augment the predictions of black-box recommender systems by interpreting feature interactions from a source model and encoding those in a target model.

**A.28.2 Code Structure, Documentation, Community, Constraints & Transductiveness.** The absence of public access to the code prevents any evaluation of its structure, documentation, community, user constraints, or transductiveness.

**A.28.3 Installation & Utilization.** Getting a working implementation is not feasible, as the method is not open-source, which is the reason for deciding the method is not usable in this context.

**A.28.4 Automation.** The method is not an AutoFE method, since it is not usable with minimal effort and without immense expertise. The required resource usage constraints can not be evaluated.

## A.29 GPFS [62]

**A.29.1 Method Description.** GPFS incorporates complex multi-variate redundant features by using genetic programming and applies state-of-the-art feature selection algorithms.

**A.29.2 Code Structure, Documentation, Community, Constraints & Transductiveness.** There is no open-source code, which means, that the code structure can not be evaluated, as well as the user constraints and the transductiveness. We can also not check if there is a documentation or a community.

**A.29.3 Installation & Utilization.** This method does not have any publicly available code, which is the reason for not further investigating and deciding it is not usable.

**A.29.4 Automation.** The method is not an AutoFE method, since it is not usable with minimal effort and without immense expertise and the required resource usage constraints are not fulfilled.

## A.30 H2O [31]

**A.30.1 Method Description.** The AutoML platform H2O automates model selection, training and ensembling focusing on generating many models in a short time. Within this platform, there exists a data preprocessing part, which is considered as a AutoFE method in this paper. The preprocessing includes encoding strategies, feature selection and feature extraction. These preprocessing parts are not further explained in the paper.

**A.30.2 Code Structure, Documentation, Community, Constraints & Transductiveness.** H2O is a large open-source project (H2O - GitHub) with maintained code form 2024, that is not well structured. There is no data publicly available. The authors provide an extensive documentation (H2O - Documentation) and there is a community of 179 contributors. As constraints, the user can specify operations to be applied on the columns of the dataset. The number and type of operations directly influences the execution time of the AutoFE method. H2O does not prevent transductive learning.

**A.30.3 Installation & Utilization.** The framework can be installed via PyPI . In order to use the H2O AutoFE part, the user needs to transform the pandas "DataFrame" to a "H2OFrame" and define a "H2OAssembly", fit it on the training data and transform the dataset by manually defining transformations, to be applied on the columns. In this process, there are occurring two errors:
```
1. ValueError: could not convert string to float
```
The value error occuring with categorical columns can be fixed with factorization. Including that, the H2O feature engineering is working, but the AutoML-Toolkit Pipeline has problems with the H2O results.
```
2. ValueError: Input X contains infinity or a value too large
for dtype('float64')
```
This can be resolved by replacing the infinite values. With these two adaptations, the feature engineering part of H2O is usable as a AutoFE method and is usable.

**A.30.4 Automation.** The method is not an AutoFE method, since it is not usable with minimal effort and without immense expertise and the required resource usage constraints are not fulfilled.

## A.31 LassoFS [63]

**A.31.1 Method Description.** The idea of using the LASSO feature selection property in order to do feature selection is described in the approach of LassoFS.

**A.31.2 Code Structure, Documentation, Community, Constraints & Transductiveness.** The lack of public access to the code makes it impossible to evaluate its structure, documentation, community, user constraints, or transductiveness.

**A.31.3 Installation & Utilization.** Since the method is not open-source, it is not further investigated and not usable.

**A.31.4 Automation.** The method is not an AutoFE method, since it is not usable with minimal effort and without immense expertise. The required resource usage constraints can not be evaluated.

## A.32 LAMA [42]

**A.32.1 Method Description.** LAMA is a AutoML framework for regression and different types of classification. The framework has a data preprocessing step included in the pipeline, which consists of two optional feature selection and one optional feature engineering step. LAMA provides different options for feature selection: No selection, Importance cut off selection (by default) and Importance based forward selection. In the scope of this paper, only the AutoFE part of LAMA is used.

**A.32.2 Code Structure, Documentation, Community, Constraints & Transductiveness.** The code from 2025 for LAMA is open-source (New LAMA - GitHub, Old LAMA - GitHub), well structured and provides example data. There is an absence of comprehensive documentation; the material provided consists solely of tutorials (LAMA - Documentation). The community consists of 15 contributors and 178 users. Users can not make any constraints to the method. LAMA prevents transductive learning.

**A.32.3 Installation & Utilization.** LAMA can be installed with PyPI . In the installation process we encountered problems with the building of a wheel. These issues cannot be easily solved, the proposed solution is to replicate the necessary components of the code for the feature engineering, specifically the "LGBSimpleFeatures" class and all its dependent code. In order to use the feature engineering, the user has to provide the data, the feature names and the roles of the features, which is the data type. The data is in the next step transformed from a pandas "DataFrame" to a "PandasDataset" and from that to a "LAMLDataset". With the data in that form, the "LGBSimpleFeatures" class can be instanced, fitted on the training data and used to transform the training and test data. While including this code, we encountered some problems:

`1. importlib.metadata.PackageNotFoundError`
The first error is a package not found error occurring in the "lightautoml/__init__.py" file where the "__version__" variable is set by using this function. It was not possible to fix this issue by getting the required package, but with debugging it is possible to determine the desired version and hard-code it instead of retrieving it dynamically.

`2. NotImplementedError: Column Slice not Implemented.`
The "not implemented"-error is associated with the data types utilized in the software. In this context, pandas "DataFrames" are used and they do not have a method for slicing the columns, but the feature engineering of LAMA uses "LAMLDataset" data types, which support this operation. The conversion from pandas "DataFrame" to a "LAMLDataset" is not trivial, as the method does not provide any functions to do so. There also exists a "PandasDataset" class, which also comes to an error when converting a pandas "DataFrame" to this class.

`3. KeyError: 'Sex'`
This specific error is due to the column name of the first column of the churn dataset from the OpenML Benchmark [[72]]. The examples for initializing a "PandasDataset" instance hard code the column names and the roles of each column, which requires to know if the column is a numerical or a categorical feature. Given that the pandas "DataFrame" is a widely used data type in machine learning, this issue should not occur within an AutoFE method, where the transformation is expected to function seamlessly. Moreover, it is impractical to hard-code the column names and roles for every dataset intended for testing; however, it is straightforward to implement code that can automatically derive this information. After doing this, there is an attribute error.

`4. AttributeError: 'DataFrame' object has no attribute 'features'`
The attribute error occurs when passing the pandas "DataFrame" or the "PandasDataset" to a "LAMLDataset" to the "fit_transform" function of the instance of "LGBSimpleFeatures", even if features are provided in the initialization of the instances and there are no errors. The effort for getting the method to work is too high, especially because the documentation can not help here, which is why it is not usable.

**A.32.4 Automation.** The method is not an AutoFE method, since it is not usable with minimal effort and without immense expertise and the required resource usage constraints are not fulfilled.

## A.33 LFS [64]

### A.33.1 Method Description.
LFS is a method for feature selection in the context of network traffic analysis. The method uses multiple learning automata techniques to determine the influence of features and select the significant ones.

### A.33.2 Code Structure, Documentation, Community, Constraints & Transductiveness.
There is no open-source code, which means, that the code structure can not be evaluated, as well as the user constraints and the transductiveness. We can also not check if there is a documentation or a community.

### A.33.3 Installation & Utilization.
This method does not have any publicly available code, which is the reason for not further investigating it. For that reason, it is also not usable.

### A.33.4 Automation.
The method is not an AutoFE method, since it is not usable with minimal effort and without immense expertise and the required resource usage constraints are not fulfilled.

## A.34 LPFS [43]

### A.34.1 Method Description.
The LPFS approach is introducing a smoothed $l_0$ function in order to select features and proposes to use this function as a gate at the input of a neural network. After the training, some features gates are 0 while others are almost one, so there is a clear distinction which features are to select.

### A.34.2 Code Structure, Documentation, Community, Constraints & Transductiveness.
There is no official GitHub of the authors of the paper, but there is an unofficial GitHub from 2022, with which it it possible to check if the method is correctly implemented and if it can be used in this context (Unofficial GitHub). The code does not follow the expected structure and there is no data example, no community and no documentation. The user can restrict the number of epochs and set the configuration of the multi-layer perceptron in order to constraint the method. LPFS is not transductive.

### A.34.3 Installation & Utilization.
LPFS is not suited for our purpose, because it needs feedback from the model for feature selection. The information on how well the features perform is in our case only available in the end of the process, which makes it impossible to feed this information back and improve the feature engineering afterwards. This method is not suited for our use case, which is the reason for deciding it is not usable.

### A.34.4 Automation.
The method is not an AutoFE method, since it is not usable with minimal effort and without immense expertise and the required resource usage constraints are not fulfilled.

## A.35 MACFE [32]

### A.35.1 Method Description.
The meta-learning and causality-based AutoFE framework (MACFE) uses meta-learning for searching optimal transformation, feature distribution encoding and causality feature selection.

### A.35.2 Code Structure, Documentation, Community, Constraints & Transductiveness.
MACFE is a open-source project from 2022(MACFE - GitHub), which is structured and does not provide any data. There is no documentation provided with the code and no active community. The user can set the two non-documented variables s_list and d_list in order to restrict the algorithm. The method does not prevent transductive learning.

A.35.3 **Installation & Utilization**. The method does not come with an installable package, the user has to use the code of the experiments.py files and all necessary dependencies from the open-source code. In our approach, the user has to define a threshold "s_list" for the feature selection part and a variable "d_list" for the feature generation part. For the feature selection, an instance of the "DAGClassifier" is fitted on the training data and transforms the data afterwards. The feature generation step adds features to the original data frame. When including this procedure in our context, one error occurred:

```
1. FileNotFoundError: [Errno 2] No such file or directory
```
In order to fix this error, it is necessary to copy several ".pkl" files from the repository. With adjusting the path, the method is usable.

A.35.4 **Automation**. The method is not an AutoFE method, since it is not usable with minimal effort and without immense expertise and the required resource usage constraints are not fulfilled.

## A.36 MAFESE [33]

A.36.1 **Method Description**. The feature selection python library MAFESE provides meta-heuristic algorithms within feature selection techniques as unsupervised-, filter-, wrapper- or embedded-based methods for finding the best attributes in high-dimensional datasets.

A.36.2 **Code Structure, Documentation, Community, Constraints & Transductiveness**. MAFESE is an open-source project (MAFESE - GitHub) last updated 2025 with a good code structure and public data. There is a documentation provided (MAFESE - Documentation), but the project does not have a community besides the three authors of the software repository. The user can restrict the AutoFE method by setting a number of features. MAFESE is not transductive.

A.36.3 **Installation & Utilization**. The method can be used with an installation coming from PyPI and is usable by fitting the "UnsupervisedSelector" instance on the training data and transforming the training and test data in the following. The inclusion of this code leads to one issue:

```
1. ValueError: could not convert string to float
```
The sole issue arises from the use of "numpy.ndarrays" in place of pandas "DataFrames", resulting in a value error when the data is not transformed correctly. By fixing the transformation, MAFESE is working and usable.

A.36.4 **Automation**. The method is not an AutoFE method, since the required resource usage constraints are not fulfilled.

## A.37 MARLFS [15]

A.37.1 **Method Description**. The multi-agent reinforcement learning framework for feature selection, MARLFS, sees each feature as an agent. The state is determined from statistical informations, an autoencoder or a graph convolution network.

A.37.2 **Code Structure, Documentation, Community, Constraints & Transductiveness**. The unavailability of public code hinders the evaluation of its structure, documentation, community, user constraints, and transductiveness.

A.37.3 **Installation & Utilization**. Getting a working implementation of MARLFS is not feasible, as the method is not open-source. For this reason, it is not usable.

A.37.4 **Automation**. The method is not an AutoFE method, since it is not usable with minimal effort and without immense expertise. The required resource usage constraints can not be evaluated.

## A.38 MEFS [65]

A.38.1 **Method Description**. MEFS uses meta-learning and ensembling to choose the best algorithm for feature selection.

**A.38.2 Code Structure, Documentation, Community, Constraints & Transductiveness.** There is no open-source code, which means, that the code structure can not be evaluated, as well as the user constraints and the transductiveness. We can also not check if there is a documentation or a community.

**A.38.3 Installation & Utilization.** Since the method is not open-source, it is not further investigated and not usable.

**A.38.4 Automation.** The method is not an AutoFE method, since it is not usable with minimal effort and without immense expertise. The required resource usage constraints can not be evaluated.

## A.39 MeLiF [66]

**A.39.1 Method Description.** The MeLiF algorithm tackles the problem of feature selection by ensembling feature ranking filters. By using a linear form coefficients space, the filter aggregation problem becomes a linear form optimization problem, which increases the speed of the feature selection.

**A.39.2 Code Structure, Documentation, Community, Constraints & Transductiveness.** The absence of public access to the code prevents any evaluation of its structure, documentation, community, user constraints, or transductiveness.

**A.39.3 Installation & Utilization.** Getting a working implementation is not feasible, as the method is not open-source, which is why it is not usable.

**A.39.4 Automation.** The method is not an AutoFE method, since it is not usable with minimal effort and without immense expertise. The required resource usage constraints can not be evaluated.

## A.40 MFE [44]

**A.40.1 Method Description.** The package MFE is a library for extracting meta-features using the most common characterization measures.

**A.40.2 Code Structure, Documentation, Community, Constraints & Transductiveness.** The authors provide open-source code form 2022 in python (MFE - GitHub (python))and R (MFE - GitHub (R)) that is well structured. There is one example dataset publicly available. MFE has a documentation. There is no community besides the 3 (in the case of the R code) or 5 (in the case of the python repository) developers. The user can not make any constraints to the method. MFE prevents transductive learning.

**A.40.3 Installation & Utilization.** The meta-features extractor is not installable via PyPI, but. The python code is easy to use and well documented, the user can fit an instance of the "MFE" class to the data and extract meta-feature names from it. The problem when including MFE is, that it is only extracting the meta-features describing a dataset, not doing feature engineering. Because of this, MFE is not usable in this context.

**A.40.4 Automation.** The method is not an AutoFE method, since it is not usable with minimal effort and without immense expertise and the required resource usage constraints are not fulfilled.

## A.41 MFS [67]

**A.41.1 Method Description.** The meta-feature learning framework aims to get the characteristics of the dataset on the basis of meta-features and chooses the best feature selection method using this information.

**A.41.2 Code Structure, Documentation, Community, Constraints & Transductiveness.** The unavailability of the code publicly hinders the evaluation of its structure, documentation, community, user constraints, and transductiveness.

**A.41.3 Installation & Utilization.** This method does not have any publicly available code, which is the reason for not further investigating and deciding it is not usable.

**A.41.4 Automation.** The method is not an AutoFE method, since it is not usable with minimal effort and without immense expertise and the required resource usage constraints are not fulfilled.

## A.42 MLJAR [34]

**A.42.1 Method Description.** MLJAR is a AutoML framework with an AutoFE part, that is used in this context. The AutoFE part of the framework creates all unique pairs of original features and derives new features by applying subtraction or division, calculates a logloss score and adds the features with the smallest score to the dataset [[73]].

**A.42.2 Code Structure, Documentation, Community, Constraints & Transductiveness.** MLJAR is open-source (MLJAR - GitHub) and last maintained 2025, well structured and provides examples. There exists an extensive documentation (MLJAR - Documentation). A community comprising 24 contributors and 127 users has been established. The user can restrict the AutoFE method by setting the number of features. Transductive learning is prevented by the method.

**A.42.3 Installation & Utilization.** The framework is installable via PyPI and is easy to use. The user needs to create a feature list, in case a feature generation is needed, otherwise, the "GoldenFeatureTransformer" can be fitted on the training data. After that, the features are selected.
```
1.  ValueError:  could not convert string to float
```
The only problem with using MLJAR is the occurrence of categorical values. By adding factorization, this problem can be fixed and the method is usable.

**A.42.4 Automation.** The method is not an AutoFE method, since it is not usable with minimal effort and without immense expertise and the required resource usage constraints are not fulfilled.

## A.43 NewApproachFS [45]

**A.43.1 Method Description.** The "New Approach for Automated Feature Selection" adds a stopping criterion to the Joint Mutual Information score and changes it by that to the so called Historical Joint Mutual Information score, which can be used for feature selection without having any actual learning.

**A.43.2 Code Structure, Documentation, Community, Constraints & Transductiveness.** The code of NewApproachFS is from 2022 and publicly available on GitHub (NewApproachFS - GitHub), not structured in the expected way and there are no examples for the data used. The authors do not provide a documentation. There is no community besides the two authors. The user can constraint the method by providing a list of expected features. NewApproachFS prevents transductive learning.

**A.43.3 Installation & Utilization.** In the method, the pymit library's "I"-function is used, that calculates the mutual information of two features. This information is then used for feature selection. The library has two problems, which make it not usable in the scope of this paper: At first, the documentation of the software is not existent and the code is hard to get. As an example, there are variable names like "X_k", "jmi", "jmi_1", "jmi_2" or "j_h", which are not understandable. In addition to that, the authors of the code are referencing code of a C++ file, which is included in the repository. It was not possible to import the file, which makes a reimplementation necessary. This is not feasible in the scope of this paper, which is why this method is not usable in this context.

**A.43.4 Automation.** The method is not an AutoFE method, since it is not usable with minimal effort and without immense expertise and the required resource usage constraints are not fulfilled.

## A.44 NFS [46]

**A.44.1 Method Description.** NFS proposes a neural architecture for AutoFE based on recurrent neural networks. A reinforcement learning-trained controller maps raw features to new features by using transformations.

**A.44.2 Code Structure, Documentation, Community, Constraints & Transductiveness.** The code from 2019 for NFS is in an unofficial version publicly available on Github (NFS - Unofficial GitHub), follows an understandable structure and provides datasets. There is no documentation and no community. The user can restrict the AutoFE method by setting a maximal order of feature operations, the batch numbers, the reinforcement learning model and the machine learning model and a number of training epochs. NFS prevents transductive learning.

**A.44.3 Installation & Utilization.** NFS can not be installed with PyPI. The method needs the instance of the "Controller" that is passed to a train function. In order to use the code, the "python-weka-wrapper" is needed, a package for using the Java Development Kit (JDK) in a python context.
```
1. ConnectionRefusedError: [Errno 61] Connection refused
```
After the installation, a connection to a locally hosted instance of "RPyC" is tried to be establish, which does not exist. There is no documentation from NFS to help with the configuration of "RPyC" or explain its purpose. The package environment can be switched from "Weka" to "Scikit-learn", which eliminates the need for "RPyC" and results in the following error message:
```
2. AttributeError: module 'tensorflow' has no attribute 'contrib'
```
"contrib" is an old tensorflow attribute that does not exist since the second version. Using a downgraded version of tensorflow is not resolving the problem. By using "tf.keras.layers.LSTMCell" instead of "tf.contrib.rnn.BasicLSTMCell", the problem is solved.
```
4. ValueError: Argument(s) not recognized: 'lr': 0.001
```
This issue can be fixed by using the proper argument name for using the "ada"m optimizer of "tensorflow", which is "learning_rate" instead of "lr".
```
5. RuntimeError: tf.placeholder() is not compatible with eager execution.
```
The runtime error occurs when using the "tf.combat.v1" version of "tensorflow". There are two possibilites: Using the old or the new version: With the new version we get an attribute error, which is related to the usage of the outdated "tensorflow" version, in this context, the second "tensorflow" version is used and there is no more "zero_state"-attribute for a "LSTMCell". Both issues are not fixable in the scope of this paper, which is the reason for not using NFS.

**A.44.4 Automation.** The method is not an AutoFE method, since it is not usable with minimal effort and without immense expertise and the required resource usage constraints are not fulfilled.

## A.45 OpenFE [12]

**A.45.1 Method Description.** The authors of OpenFE introduce a boosting method for the performance estimation of generated features and a pruning algorithm in two steps for selecting the most promising candidate features out of the generated features.

**A.45.2 Code Structure, Documentation, Community, Constraints & Transductiveness.** The AutoFE method OpenFE is open-source (OpenFE - GitHub) with well structured code and an example dataset. It was last maintained may 2024. The authors of OpenFE provide a small documentation, that only contains installation informations and a small example (OpenFE - Documentation). This is not a proper documentation. There is no community besides the two authors of the software. The user can not restrict OpenFE with any constraints. Transductive learning is not prevented.

A.45.3 **Installation & Utilization.** OpenFE is installable with PyPI and easy to use. An instance of "OpenFE" needs to be created, which can be fitted on the dataset and automatically adds selected generated features to the dataset. One problem with OpenFE is, that it can not be parallelized trivially, since it is reading and writing in a file, which leads to conflicts in parallel execution. This can be solved by giving more specific names to the files and write by that in different files for different parallelized runs. Other than that OpenFE has no problems and for that usable.

A.45.4 **Automation.** The method is not an AutoFE method, since the required resource usage constraints are not fulfilled.

## A.46 OptFS [47]

A.46.1 **Method Description.** In OptFS, each feature interaction selection is dissolved into the selection of two correlated features.

A.46.2 **Code Structure, Documentation, Community, Constraints & Transductiveness.** OptFS is a open-source AutoFE method (OptFS - GitHub) from 2022, which is well structured and provides three datasets. There exists no documentation for the code and there is no community. The user can constraint the AutoFE method by setting the number of features, the number of maximal epochs and the embedding dimensions of the neural network. The method prevents transductive learning.

A.46.3 **Installation & Utilization.** The method is not installable via PyPI. It is using three datasets in order to show the purpose of the method. When looking in the code, the feature names are hard coded. Since the considered methods are supposed to be automatic and we do not want to write feature names and parameters like "latent_dim", "feat_num" or "field_num" in our code, this method is not usable without a coding effort. This implementation effort is not feasible within the scope of this paper, which is why this method is not usable.

A.46.4 **Automation.** The method is not an AutoFE method, since it is not usable with minimal effort and without immense expertise and the required resource usage constraints are not fulfilled.

## A.47 RegularizedFS [68]

A.47.1 **Method Description.** RegularizedFS uses regularization for selecting embedded features in bug prediction models.

A.47.2 **Code Structure, Documentation, Community, Constraints & Transductiveness.** The public unavailability of the code hinders the evaluation of its structure, documentation, community, user constraints, and transductiveness.

A.47.3 **Installation & Utilization.** Getting a working implementation is not feasible, as the method is not open-source. This is the reason for deciding the method is not usable.

A.47.4 **Automation.** The method is not an AutoFE method, since it is not usable with minimal effort and without immense expertise. The required resource usage constraints can not be evaluated.

## A.48 Relief [69]

A.48.1 **Method Description.** Relief-based algorithms are a subgroup of filter-style algorithms that can be used for feature selection. Relief and ReliefF are the two relief-based algorithms that consider feature interactions while ignoring correlations in between features.

A.48.2 **Code Structure, Documentation, Community, Constraints & Transductiveness.** There is no open-source code, which means, that the code structure can not be evaluated, as well as the user constraints and the transductiveness. We can also not check if there is a documentation or a community.

**A.48.3 Installation & Utilization.** Since the method is not open-source, it is not further investigated and not usable.

**A.48.4 Automation.** The method is not an AutoFE method, since it is not usable with minimal effort and without immense expertise. The required resource usage constraints can not be evaluated.

## A.49 ReinSel [70]

**A.49.1 Method Description.** The approach of ReinSel uses existing feature selection methods in order to augment diversity in Classifier Ensembles. The proposed method aims on evaluating methods with a reinforcement learning-based method.

**A.49.2 Code Structure, Documentation, Community, Constraints & Transductiveness.** The lack of publicly available code makes it impossible to check its structure, documentation, community, user constraints, or transductiveness.

**A.49.3 Installation & Utilization.** Since the method is not open-source, it is not further investigated and not usable.

**A.49.4 Automation.** The method is not an AutoFE method, since it is not usable with minimal effort and without immense expertise. The required resource usage constraints can not be evaluated.

## A.50 RLFS [16]

**A.50.1 Method Description.** RLFS follows the idea of restricting the sate space and by that simplifying the selection of features. The authors propose to use a single agent traversing the environment in a reinforcement learning setting. A SVM classifier is evaluating selected features, and its output is used as a reward for the agent.

**A.50.2 Code Structure, Documentation, Community, Constraints & Transductiveness.** The absence of public access to the code prevents any evaluation of its structure, documentation, community, user constraints, or transductiveness.

**A.50.3 Installation & Utilization.** This method does not have any publicly available code, which is the reason for not further investigating it. For that reason, it is not usable.

**A.50.4 Automation.** The method is not an AutoFE method, since it is not usable with minimal effort and without immense expertise. The required resource usage constraints can not be evaluated.

## A.51 SARLFS [17]

**A.51.1 Method Description.** SARLFS proposes a single-agent reinforced feature selection including a restructured choice strategy. The strategy consist of exploiting one agent for the selection task of several features and introducing a scanning method, which encourages the agent to treat more than one features in each scanning round.

**A.51.2 Code Structure, Documentation, Community, Constraints & Transductiveness.** There is no open-source code, which means, that the code structure can not be evaluated, as well as the user constraints and the transductiveness. We can also not check if there is a documentation or a community.

**A.51.3 Installation & Utilization.** Getting a working implementation is not feasible, as the method is not open-source, which is why it is not usable.

**A.51.4 Automation.** The method is not an AutoFE method, since it is not usable with minimal effort and without immense expertise. The required resource usage constraints can not be evaluated.

### A.52 SimplerIsBetter [48]

**A.52.1 Method Description.** The SimplerIsBetter framework uses black-box models in order to use them for generating simpler white-box models, which are created with engineered features [].

**A.52.2 Code Structure, Documentation, Community, Constraints & Transductiveness.** The code of this method is from 2021 and publicly available (SimplerIsBetter - GitHub), well structured and does not contain any data. There is a documentation in the GitHub repostitory, which is only contains the abstract of the paper and can for that reason not be counted as documentation. There is no community around this project besides the two authors of the paper. The user can not make any constraints to the method. SimplerIsBetter prevents transductive learning.

**A.52.3 Installation & Utilization.** The programming language used for implementing this open-source method is R, which is not easily transferable to the one used in this context, which is python. This makes it necessary to reimplement the method from scratch, which is not feasible.

**A.52.4 Automation.** The method is not an AutoFE method, since it is not usable with minimal effort and without immense expertise and the required resource usage constraints are not fulfilled.

### A.53 SVMFS [71]

**A.53.1 Method Description.** This wrapper algorithm for selecting features is based on support vector machines. SVMFS uses sequential backward selection, where the error count of the validation data determines which features are to remove.

**A.53.2 Code Structure, Documentation, Community, Constraints & Transductiveness.** The lack of public access to the code makes it impossible to evaluate its structure, documentation, community, user constraints, or transductiveness.

**A.53.3 Installation & Utilization.** Since the method is not open-source, it is not further investigated and not usable.

**A.53.4 Automation.** The method is not an AutoFE method, since it is not usable with minimal effort and without immense expertise. The required resource usage constraints can not be evaluated.